\documentclass[letterpaper, 10 pt, conference]{ieeeconf}  

\IEEEoverridecommandlockouts                              

\usepackage{graphics} 
\usepackage{epsfig} 
\usepackage{mathptmx} 
\usepackage{times} 
\usepackage{amsmath} 
\usepackage{amssymb}  
\usepackage[ruled,vlined]{algorithm2e}
\usepackage{multirow}
\usepackage{color}
\usepackage{url}
\usepackage{hyperref}
\usepackage{subfigure}
\usepackage{booktabs}
\usepackage{algpseudocode}

\title{\LARGE \bf
 Towards Hierarchical Task Decomposition\\ using Deep Reinforcement Learning for Pick and Place Subtasks}

\author{Luca Marzari$^{1,+}$,  Ameya Pore$^{1,2}$, Diego Dall'Alba$^{1}$, Gerardo Aragon-Camarasa$^{3}$,\\ Alessandro Farinelli$^{1}$ and Paolo Fiorini$^{1}$
\thanks{$^{1}$ Department of Computer Science, University of Verona, Verona, Italy
        {\tt\small }}%
\thanks{$^{2}$ Center of Research in Biomedical Engineering, Universitat Politècnica de Catalunya , Barcelona, Spain
        {\tt\small}}%
\thanks{$^{3}$ Computer Vision and Autonomous group, School of
Computing Science, University of Glasgow, Glasgow, UK
        {\tt\small}}%
\thanks{$^{+}$ corresponding author: 
        {\tt\small luca.marzari@studenti.univr.it}}%
}

\begin{document}

\maketitle
\thispagestyle{empty}
\pagestyle{plain}

\begin{abstract}
Deep Reinforcement Learning (DRL) is emerging as a promising approach to generate adaptive behaviors for robotic platforms.
However, a major drawback of using DRL is the data-hungry training regime that requires millions of trial and error attempts, which is impractical when running experiments on robotic systems. 
Learning from Demonstrations (LfD) has been introduced to solve this issue by cloning the behavior of expert demonstrations. However, LfD requires a large number of demonstrations that are difficult to be acquired since dedicated complex setups are required.
To overcome these limitations, we propose a multi-subtask reinforcement learning methodology where complex pick and place tasks can be decomposed into low-level subtasks. These subtasks are parametrized as expert networks and learned via DRL methods. 
Trained subtasks are then combined by a high-level choreographer to accomplish the intended pick and place task considering different initial configurations. As a testbed, we use a pick and place robotic simulator to demonstrate our methodology and show that our method outperforms a benchmark methodology based on LfD in terms of sample-efficiency.
We transfer the learned policy to the real robotic system and demonstrate robust grasping using various geometric-shaped objects. 
\end{abstract}

\section{Introduction}
Robot learning has been an emerging paradigm since the advent of Deep Reinforcement Learning (DRL) with breakthroughs in dexterous manipulation \cite{levine2018learning}, grasping \cite{gu2017deep} and navigation for locomotion tasks \cite{haarnoja2018learning}.
However, a significant barrier in the universal adoption of DRL for robotics is the data-hungry training regime that requires millions of trial and error attempts to learn goal-directed behaviors, which is impractical in real robotic hardware. Existing DRL methods learn complex tasks end-to-end, leading to overfitting of training idiosyncrasies, which makes them sample inefficient and less adaptable to other tasks \cite{zhang2018dissection}.
Therefore, new DRL policies have to be trained from scratch even for solving problems that are highly similar to the pretrained task, which leads to wastage of computation power.
\begin{figure}[thpb]
      \centering
      \includegraphics[width=0.43\textwidth]{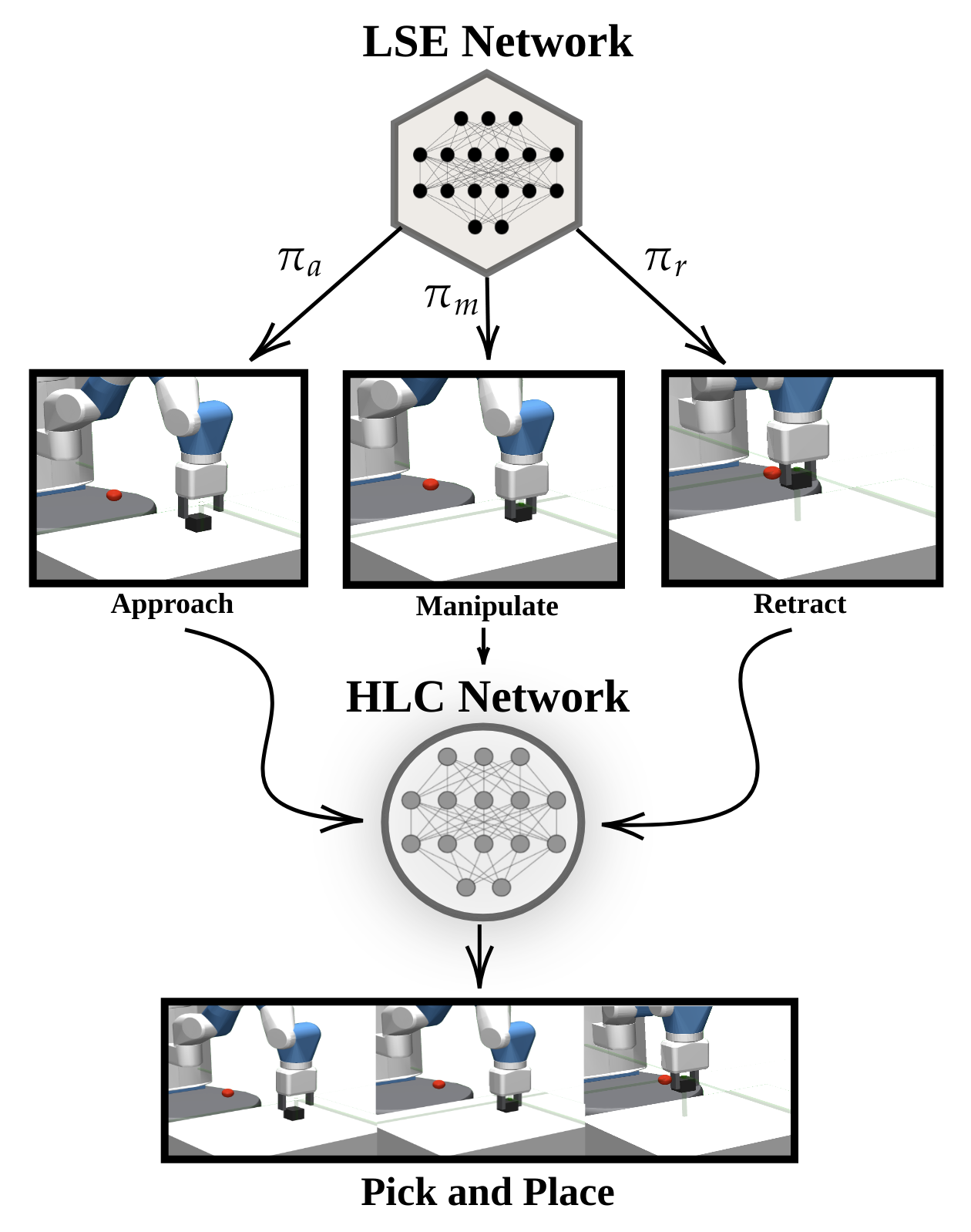}
      \caption{Summary diagram of the hierarchical architecture proposed in this paper. The pick and place task is divided into Low-level Subtask Experts (LSE), namely \textit{approach}, \textit{manipulate} and \textit{retract}. These subtasks are coordinated using a High Level Choreographer (HLC).}
      \label{general_arch}
      \vspace{-5mm}
 \end{figure}

Learning from Demonstrations (LfD) approaches have been designed to be efficient with respect to end-to-end DRL methods, since they train a neural network to clone the expert behavior described by a dataset of reference trajectories.
However, such techniques require a considerable number of demonstrations to be trained adequately, in addition to specialized data-acquisition hardware and instrumentation, such as virtual reality or teleoperation units \cite{pore2021learning}. Moreover, the LfD approach limits robot performance as it can only be as good as a reference trajectory since there is no additional feedback for improvement. Also, commonly used LfD techniques such as Behavior cloning (BC) suffer from compounding errors in long time horizon tasks \cite{ross2011reduction}.

An alternative approach to LfD is based on modularization of a neural network to encode certain attributes of a complicated control problem \cite{yang2020multi, devin2017learning}. These subtask modules can be assembled in a variety of combinations to output versatile behaviors. Some advantages of using subtask networks are: 
(i) the subtasks networks are much easier and faster to train than learning an overall control policy;
(ii) modular behaviors are easier to interpret and can be adapted to similar tasks \cite{xu2019toward}. On the contrary, modular approach requires a priori knowledge about the task for designing subtask networks, however this information is much less demanding compared to expert demonstrations in LfD.
Therefore, in this paper, we hypothesize that an end-to-end complex control task can be simplified into high-level subtasks using the domain knowledge of the human operator and these subtasks can be in turn learned using a DRL method. 
DRL for low-level subtasks will ensure that the learned policy considers the environment and mechanical constraints of the robot rather than human bias from the demonstrations.
This approach is partially inspired by Hierarchical Reinforcement Learning (HRL) methods that operate multiple policies at different temporal scales. 
However, a significant drawback of these approaches is that learning multiple hierarchical policies simultaneously can be unstable \cite{nachum2018data}. To overcome this challenge, we train the low-level policies independently from the high-level policy.



In this work, we consider a robotic pick and place task and decompose the task into simpler subtasks, namely \textit{approach}, \textit{manipulate} and \textit{retract}. These subtasks are trained independently using a DRL policy with a sub-goal directed reward function for each subtask.
Further, the subtasks are coordinated by a High Level Choreographer (HLC) network that learns to sequence subtasks to output the intended behaviors (see Fig.~\ref{general_arch}). Our approach is inspired by learning patterns followed by humans while acquiring new skills. Humans learn complex tasks by segmenting them into more superficial behaviors and learning each of them separately. As an example of complex motor skills involved in basketball, the athlete learns to execute dribbling, passing and shooting during training. These low-level skills can then be combined to output complex skills such as assisting, attacking, freeball, to name a few \cite{jia2020mastering}.

Hence, our contribution is a multi-subtask DRL methodology to learn pick and place tasks. We provide a comparative analysis of our method with an established LfD baseline. Towards the end, we show the successful transfer of the learned policies to a real robotic system and measure its success in grasping different geometrically shaped objects.

The outline of this paper is as follows. In Sec.~\ref{relatedworks}, we summarize studies in the literature that are close to this work. The proposed training methodology is explained in Sec.~\ref{methods}. 
In Sec.~\ref{experiments}, we describe experiments performed to validate our approach, then the results obtained are presented in Sec.~\ref{resutls}. Finally, we conclude and provide directions towards future work in Sec.~\ref{conclusion}.

\section{Related Work} \label{relatedworks}

\textit{Learning from demonstrations (LfD)}: An alternative to make end-to-end DRL algorithms efficient and learn human-like behavior is LfD. Vecerik et al. \cite{vecerik2017leveraging} used demonstrations to fill a replay buffer to provide the agent with prior knowledge for a Deep Deterministic Policy Gradient (DDPG) policy. Nair et al. \cite{nair2018overcoming} showed that task demonstrations could be used to provide reference trajectories for DRL exploration. They introduced a Behavior Cloning (BC) loss to the DRL optimization function and showed that the agent is more efficient than a baseline DDPG method. Similarly, Goecks et al. \cite{goecks2019integrating} proposed a two-phase combination of BC and DRL, where demonstrations were used to pretrain the network followed by training a DRL agent to produce an adaptable behavior. 
In this paper, we experiment with the latter for training LSE to overcome the challenges with data-acquisition in BC. 

\textit{Task decomposition}: The idea of splitting a task into subtasks has been reported in the literature \cite{jacobs1991adaptive, brooks1991intelligence}, in which these subtasks are then choreographed to output a complex behavior.
Recently, HRL has emerged as a reinforcement learning setting where multiple agents can be trained at various levels of temporal abstraction \cite{barto2003recent} and learn different subtasks following an end-to-end training paradigm. HRL consist of training agents such that the low-level agent encodes primitive motor skills while the higher-level policy selects which low-level agents are to be used to complete a task \cite{nachum2018data, levy2017hierarchical}. Similarly, Beyret et al. \cite{beyret2019dot} studied an explainable HRL method for a robotic manipulation task that uses Hindsight Experience Replay (HER) as a high-level agent to decide goals that are given as input to the low-level policy. In these works, hierarchical policies are learned end-to-end, thus observe instability leading to sample inefficiency, i.e., the lower level policy changes under a non-stationary high-level policy. In this work, to overcome this limitation we propose to train the LSE independently from the high-level policy HLC.

\textit{Muti-subtask approaches}: Yang et al. proposed to use sets of pretrained motor skills parametrized by a deep neural network \cite{yang2020multi}. From these pretrained motor skills, a gating network learns to fuse the networks' weights to generate various multi-legged locomotion tasks. In this work, we do not use network fusion; rather we combine the subtasks using a choreographer. Devine et al. explored modular neural network policies to learn transferable skills for multi-task and multi-robot  \cite{devin2017learning}. Some recent studies have shown advances in the modularization of neural networks in which complicated control policies can be modularized as a series of attributes \cite{xu2019toward, pore2020simple}. Each of these attributes is trained separately and assembled to produce a required behavior. Xu et al. use parallel attribute networks to combine parallel skills simultaneously \cite{xu2019toward}. Whereas Pore et al. use BC to learn individual subtask networks and then combine them using a high-level DRL network \cite{pore2020simple}. 
In this work, we propose to train each individual subtask using a pure DRL approach, thus overcoming the limitations of both BC and parallel attribute networks proposed in previous works.
\begin{figure*}[h!]
\centering
\includegraphics[width=\linewidth]{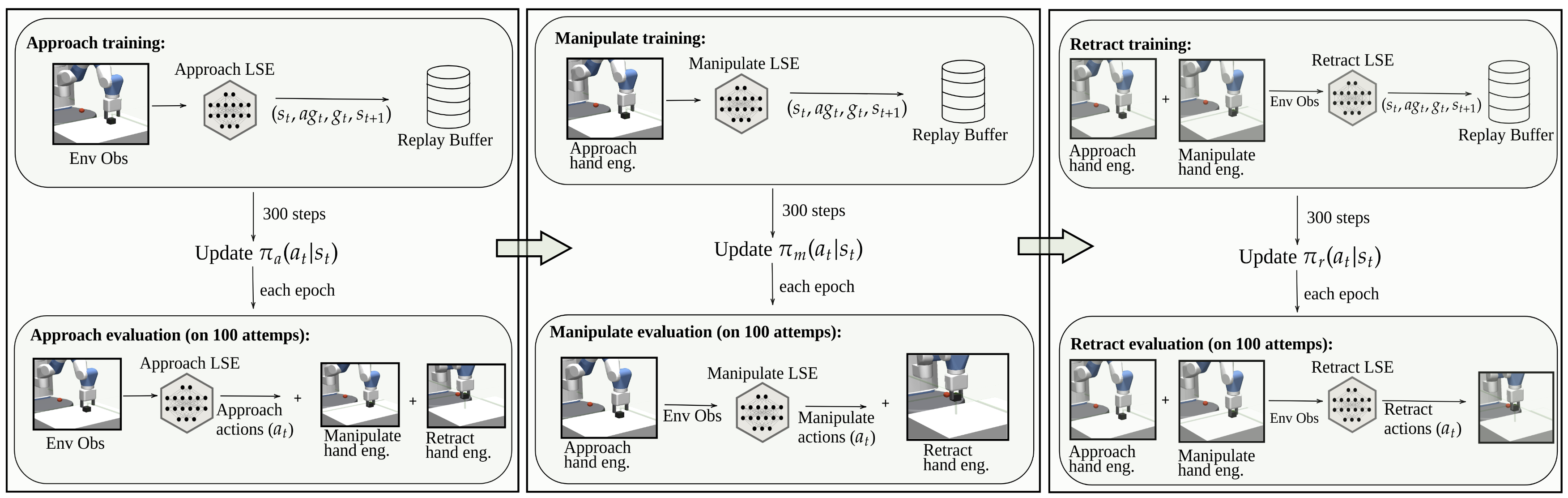}
\caption{Schematic overview of the LSE training and evaluation process: All the LSE are trained independently (from left to right) approach, manipulate and retract respectively. The LSE policy $\pi$ is updated offline by sampling from a replay buffer after 300 steps using DDPG+HER. The policy is evaluated after each epoch by using hand-engineered solution for other subtasks by computing the success rate on 100 episodes.}
\label{trainingProc}
\vspace{-4mm}
\end{figure*}

\section{Methods}\label{methods}
We consider the pick and place robotic task, where the robot's goal is to grasp a randomly positioned object in the environment within reach of the robot and place it to a target location. 
The task is manually decomposed into three subtasks: approaching the current object position, manipulating the object to grasp it, and retracting the object to a target position. For learning the entire pick and place task, we consider two Markov Decision Processes (MDPs) with different levels of temporal hierarchy. 
The higher-level agent (HLC), acts at the level of subtasks and learns a policy to choreograph the subtasks, whereas a low-level agent (LSE), learns the policy for low-level actions inside the subtask. Since a hierarchical task decomposition is used, there are two different goals during each episode: a subtask goal that is considered to train LSE and the final task goal that is used to train the HLC. In Sec.~\ref{LSE}, we describe the training strategy for the LSE agent, and in Sec.~\ref{HLC}, we describe the HLC training strategy\footnote{Project code: \url{https://github.com/LM095/DRL-for-Pick-and-Place-Task-subtasks}}.

\subsection{Training the Low-level Subtask Expert (LSE)}\label{LSE}

The aim of an LSE is to learn an optimal policy and task representations to accomplish the specific subtask. For this, we define an MDP formulation for LSE as follows. Inside each subtask $u_i$ (where $i, 1\leq i \leq3$ refers to the number of subtasks), for every time step $t$, the agent receives a state input $S_t$ from the environment $E$, executes an action $a_t$ and moves to the next state $S_{t+1}$. We use a DDPG + HER training paradigm to learn the LSE policy $\pi_{u_i}$ 
since it has been demonstrated as an efficient candidate for end-to-end pick task \cite{andrychowicz2017hindsight, multigoal2018}.
The state inputs to the agent are the vector observations that provide the kinematic information (such as position, velocity, and orientation) of the object and the robotic gripper. The action output of the LSE network consists of $x$, $y$, and $z$ positions. Each of the LSE is parametrized by a neural network that consists of three fully connected layers with ReLu activation functions and one final linear output layer with Tanh activation function in case of actor and without activation function in case of the critic. 

The training process works as follows: for each subtask, $u_i$ at each episode, i.e., 300 steps, we store a list of tuples $(s_t, ag_t, sg_t, s_{t+1})$ in the replay buffer where $s_t$ is the observation at the beginning of the episode, $ag_t$ is the achieved goal after taking action during the episode, i.e., the new gripper position, $sg_t$ is the goal of the subtask during the episode, and $s_{t+1}$ is the new state after completing the action in the environment.
We use a dense reward function $r_t$ that is defined as:
\[r_t = -d(ag_t - sg_t)\] i.e., it returns the negative Cartesian distance $d$ between the achieved goal and the subtask goal at each timestep.
We use DDPG+HER paradigm to sample state observations from the replay buffer and perform an update of  $\pi_{u_i}$ every 300 steps. Finally, after each epoch (15k steps), we evaluate the learning level of the LSE, using hand-engineered actions for the subtasks that are not being trained. Kindly refer to Fig.~\ref{trainingProc} for the schematic overview of the described method.
Hand-engineered solutions are pre-configured action values used to reach a desired target state. In the evaluation process of \textit{approach}, action output from the LSE network are used for the approach subtask, and hand-engineered actions used for \textit{manipulate} and \textit{retract}.
In this way, if at the end of the episode the block fails to be placed at the target position, it implies that the \textit{approach} part has not been successful and needs to be trained further. Note that the engineered solutions are only required to reach a intended position before training a specific LSE module and for the evaluation phase to test if the robot can complete the task successfully.
\begin{figure*}[ht]
\centering
\includegraphics[width=0.8\linewidth]{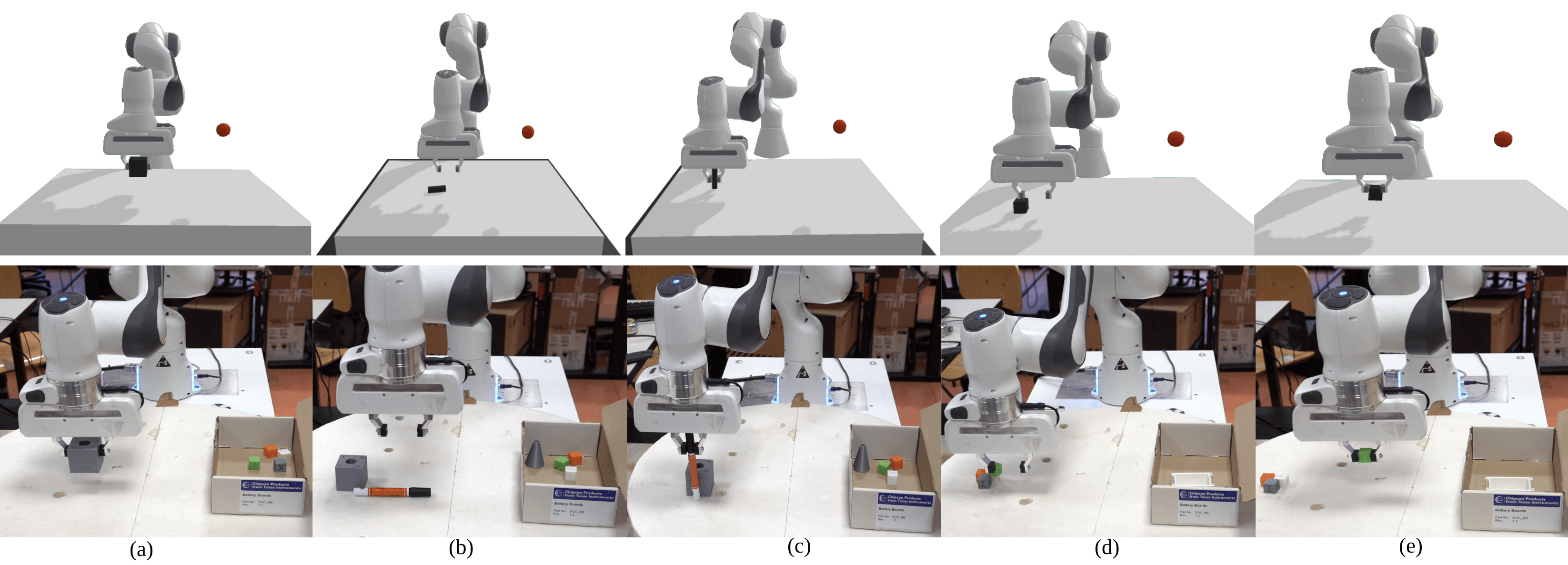}
\caption{Pick and place task (a) accomplished with end-to-end learning strategy with DDPG+HER and our LSE DDPG+HER. (b) failure with a thin cylindrical object for end-to-end strategy (c) success with a narrow cylindrical object for the agent trained with our LSE strategy. (d)  failure with a small box object for the agent trained with end-to-end strategy. (e) success with a small box object for the agent trained with our LSE strategy.
}
\label{fig:experiments}
\vspace{-4mm}
\end{figure*}
\subsection{High Level Choreographer (HLC)} \label{HLC}
After the LSE networks are trained, we establish an HLC that learns a policy to choreograph the subtasks to complete the task temporally. For an HLC agent in a state $s_{t'}$, it activates a subtask $u_i$ and receives a reward $r_{t'}$. As a consequence, the agent goes to a state $s_{t'+1}$ that corresponds to the state after completing the activated subtask. Note that the notation $t$ and $t'$ are used to indicate temporal hierarchy, i.e., $t$ refers to timestep for the LSE networks, while $t'$ refers to a timestep for the HLC. We use an actor-critic network architecture introduced in \cite{pore2020simple} where the actor policy selects one of the subtask. The network consists in the recurrent network followed by two independent, fully connected layers that serve as the actor and critic. 

Since the output of the HLC network is a discrete action value, we use an asynchronous Actor-Critic (A3C) training strategy to learn the HLC policy \cite{mnih2016asynchronous}. Further, generalized advantage estimation \cite{schulman2015high} is used to improve data efficiency and reduce the variance in the trajectories. 
We define a  \textit{sparse} reward function $r_{t'}$ where the HLC receives a positive reward if the robot is able to place the block at the target position, i.e. the HLC chooses the correct subtask sequence.

\section{Experiments} \label{experiments}
To validate our hypothesis, we perform two sets of experiments. First, a comparative study between our LSE approach and a baseline LSE trained via BC.
We use BC as a baseline as it has been demonstrated to be efficient compared to an end-to-end DRL method \cite{pore2020simple}.
Second, we show the successful translation of the learned policy from the simulator to a real robotic system. The training methods are carried out on an Intel Core i7 9th Gen system. 
\begin{figure}[h!]
\centering
\includegraphics[width=0.7\linewidth]{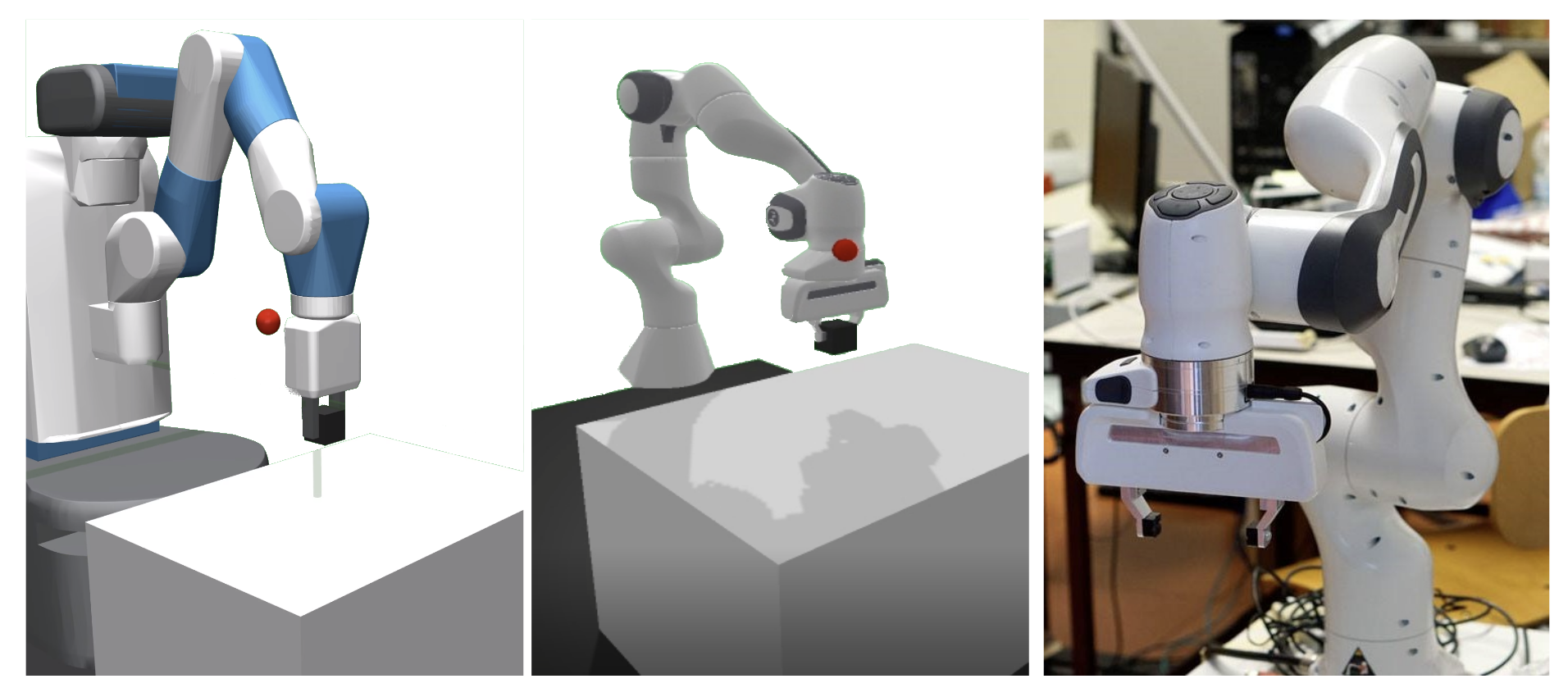}
\caption{Different environments used for experiments  (a) \textit{FetchPickAndPlace-v1} (b) \textit{PandaPickAndPlace-v0} (c) Franka Emika Robot used for real robot demonstrations.}
\label{figure3}
\end{figure} 

\subsection{Simulation experiments: \textit{FetchPickAndPlace-v1}}
In the first experiment, we use the Mujoco simulation engine environment, \textit{FetchPickAndPlace-v1} that comprises the Fetch robot (see Fig.~\ref{figure3}a). 
In order to use dense reward in each LSE, we modified the original $step$ function \cite{brockman2016openai}, which allows the robot to act in the environment given a chosen action and obtain a new state and a reward. Our new $step$ function requires two parameters: the action to be taken in the environment and the agent's goal.

Once an LSE reaches a high success rate, the weights are saved, and we use a similar strategy to train the remaining subtasks, as described in section \ref{methods}A. 
Finally, after each of the subtasks is trained, we load the network's weights and train the HLC to choreograph the subtasks temporally. For each LSE, we show the training performance using two methods trained via DDPG+HER and BC, following the schematics shown in Fig.~\ref{trainingProc}.

\subsection{Real robot experiments}
For the second part of our experiments, we transfer all the methodology presented in the section \ref{methods} to another simulation environment, called \textit{PandaPickAndPlace-v0}\footnote{\url{https://github.com/qgallouedec/panda-gym}}, that consists of the Franka Emika Robot. This step was carried out to facilitate the transfer to the real Franka Emika robot available in our laboratory. 
To replicate the challenges found in the real system, namely the the difficulty of obtaining dense rewards at every time step, we modify the observation space in \textit{PandaPickAndPlace-v0} to consider only the variables that can be measured in the real robotic system.
Hence, for both simulation (\textit{PandaPickAndPlace-v0}) and real robot, we consider the current pose of the gripper, the initial pose of the object, and the state of the joints of the gripper.

We establish the communication pipeline between the simulation environment and the real robot using a Robot Operating System (ROS) node that is interfaced with the \textit{Moveit} framework\footnote{\url{https://moveit.ros.org/}}. 
The poses generated by the actions in the \textit{PandaPickAndPlace-v0} environment are processed by \textit{Moveit} to generate the complete trajectory while observing the physical constraints of the real robot.
We apply a homogeneous transformation to change the reference frame, which lies at the gripper center in the simulation scene, to the panda base frame in the real robot.

Lastly, we reuse the subtasks and fine-tune the LSE retract to grasp different types of objects. An end-to-end learning approach would require complete retraining for different objects. 
Our objects include two different geometrical-shaped objects such as a cylinder and a block of different dimensions used in the training procedure (see Fig.~\ref{fig:experiments}). LSE approach provides a possibility to change one of the subtasks without affecting other trained subtasks. Using a subset of behaviors is not possible in end-to-end learning. Hence, in the proposed method, we use the trained LSE on the block pick and place task and fine-tune the grasping for the \textit{retract} subtask, whereas we directly deploy the behaviors learn for the end-to-end learning. 

\section{Results} \label{resutls}
\begin{figure}[h!]
\includegraphics[width=\linewidth]{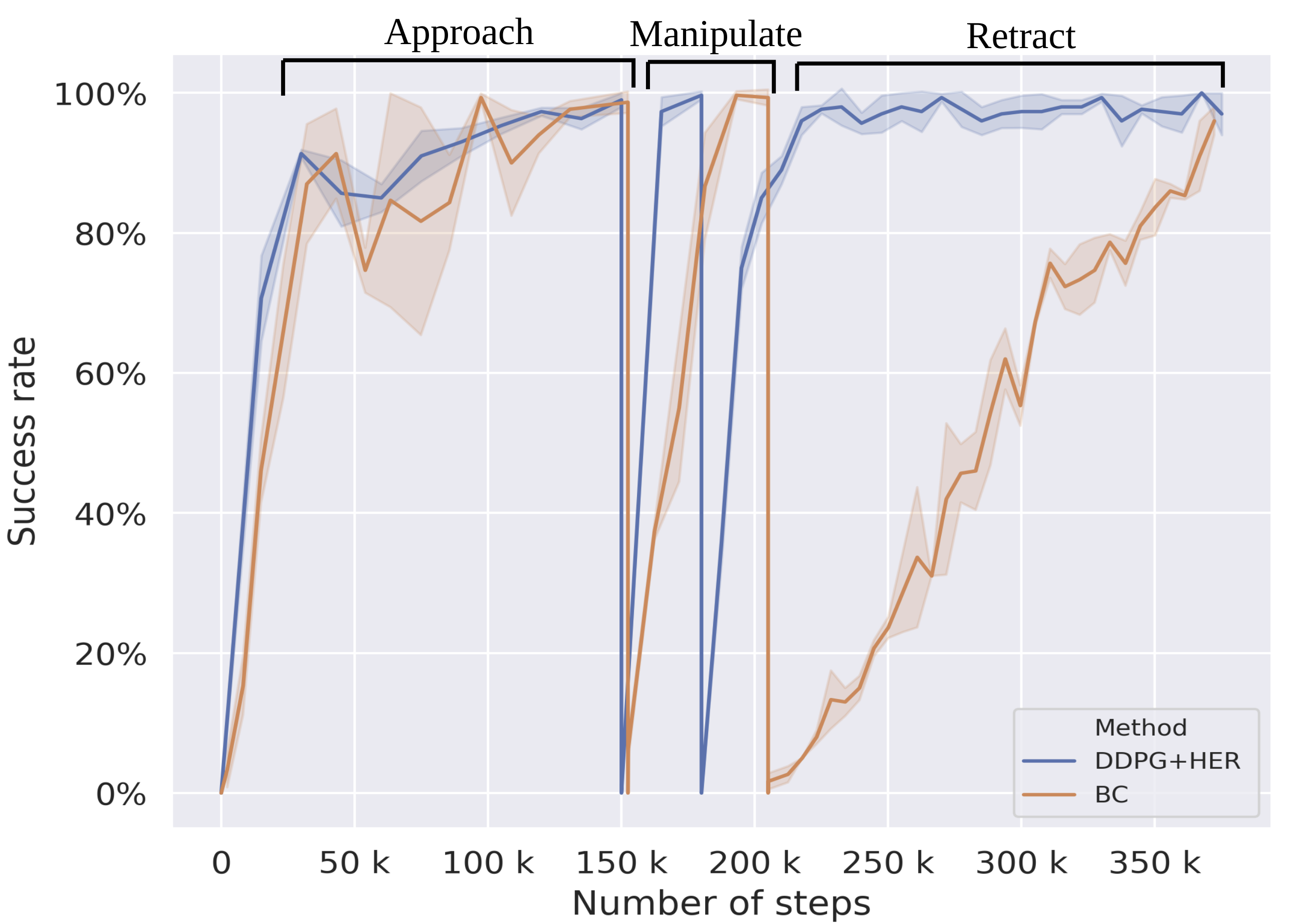}
\caption{Performance comparison of our training strategy using DDPG+HER and BC. Each experiment is executed independently three times with different seeds. Success is quantified as the percentage of successful grasp as a function of training steps.}
\label{figure5}
\end{figure}

Firstly, we provide the results for the comparative performance of our method, which uses DRL techniques for training LSE with an established LfD baseline using BC.
Fig.~\ref{figure5} depicts the sample efficiency of the LSE strategy trained via DDPG+HER and BC learning method. The peak represents the maximum success reached by each method for each subtask, i.e., the first peak denotes the completion of training the \textit{approach} subtask, the second peak denotes completion of the training of \textit{manipulate} subtask, and the third peak indicates training the \textit{retract} subtask.  DDPG+HER outperforms BC and reaches 100\% success in 218k steps, while BC takes 372k steps.

Moreover, DDPG+HER shows a smooth monotonous learning curve compared to BC, which does not stabilize immediately after reaching high success values. Overall, DDPG+HER shows less variance compared to BC. There is a significant difference between the learning curve for the \textit{retract} behavior. \textit{retract} is a temporally elongated subtask compared to other subtasks, i.e., out of 50 timesteps in an entire episode, \textit{retract} subtask takes 25 timesteps. Due to the long horizon task, BC seems to suffer from the compounding error caused by a covariate shift. Hence, we observe DDPG+HER faster in learning for the \textit{retract} subtask.


Table \ref{tab:steps} shows the comparison of training performance of the methods presented in this work. In particular, we analyze two possible strategies. 
The first strategy refers to a subtask approach using BC, and the second one refers to the new methodology proposed in this paper. For the strategies that use subtasks, we define LSE1 as $approach$, LSE2 as $manipulate$, and LSE3 as $retract$.
DDPG+HER using subtask decomposition is the best performing approach, and our results suggest that following the subtask approach, training can be more effective if we use a DRL algorithm than supervised BC. The behavior learned by DDPG+HER is more robust and does not require the collection of expert demonstrations, which can be time-consuming and often reflects less variability. Moreover, training using a subtask approach shows a significant reduction in both steps (by $\sim$ 77\%) and time ( by $\sim$ 75\%) with respect to end-to-end training and therefore is the best training strategy in this context.

\begin{table}[ht]
 \centering
 \scriptsize
\caption{Performance of methods for the same level of success rate}
 \label{tab:steps}
 \begin{tabular}{p{0.15\linewidth}cccccc}
\hline\noalign{\smallskip}
       & \multicolumn{5}{c}{Number of steps} & {Total time} \\
\noalign{\smallskip }\cline{2-6} \noalign{\smallskip} 
    &        LSE1 &          LSE2 &        LSE3 &          HLC &  Total   \\
\noalign{\smallskip}\hline\noalign{\smallskip}
DDPG+HER end-to-end & - & - & - & - & 1.4M & $\sim$1h\\
\rule{0pt}{3ex}%
BC LSE & 152k & 52k & 168k & 98k & 470k &  $\sim$25 min\\
\rule{0pt}{3ex}%
DDPG+HER LSE & 150k & 30k & \textbf{38k} & 98k & \textbf{316k} &  $\sim$18 min \\
\noalign{\smallskip}\hline
\end{tabular}
\vspace{-3mm}
\end{table}

\begin{figure*}[h]
\centering
\includegraphics[width=0.8\linewidth]{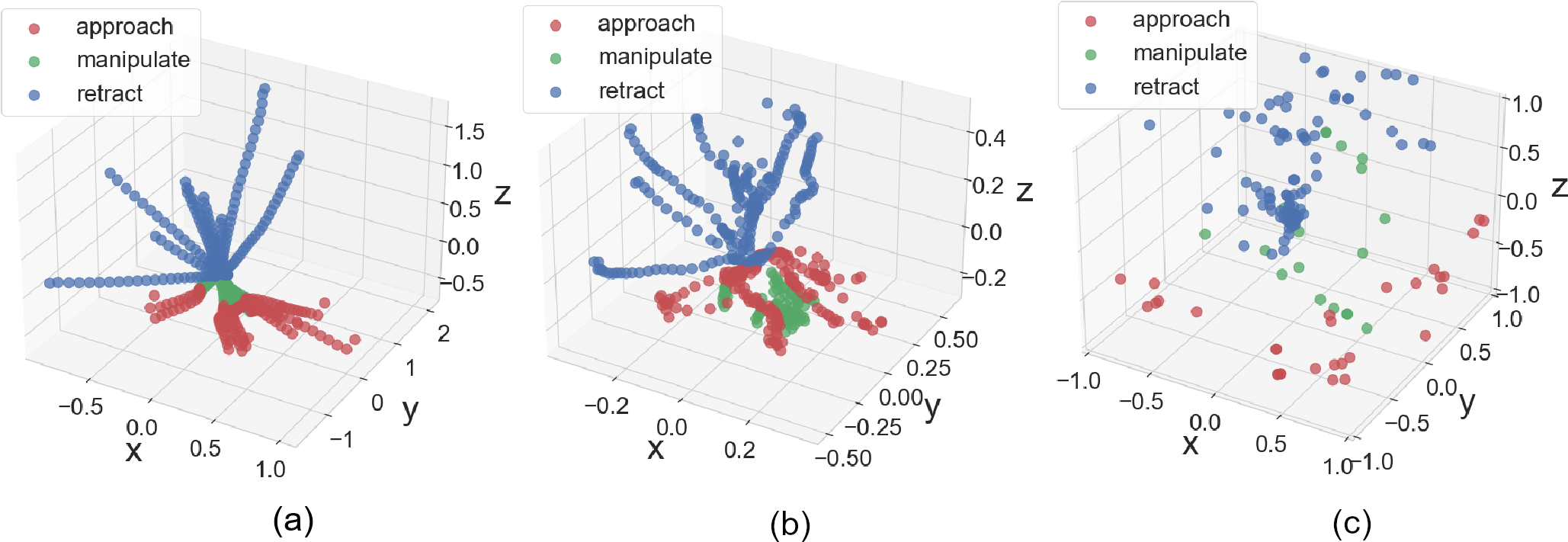}
\caption{LSE specialization analysis using different training strategies. Samples representing activation patterns using (a) hand-engineered solutions (b) learned using our subtask approach (c) learned using an end-to-end strategy for ten episodes.}
\label{fig:characterisation}
\vspace{-4mm}
\end{figure*}
 
We analyze the actions learned by LSE policy and an end-to-end policy in Fig.~\ref{fig:characterisation}. For this, we take trained LSEs (i.e., 100\% success rate for each subtask) and the end-to-end model, respectively, and analyze their activation patterns in the Cartesian space for ten episodes. Note that the initial environment conditions are the same for both policies. Fig.~\ref{fig:characterisation}a shows the specialized subtask activation patterns of ground truth hand-engineered solutions. Fig.~\ref{fig:characterisation}b and ~\ref{fig:characterisation}c depict the actions learned using our proposed approach and end-to-end learning strategy, respectively. The actions generated from LSE networks are in the vicinity of the hand-engineered actions (see Fig.~\ref{fig:characterisation}a), indicating that the learned behavior is specialized to the particular subtask. There is a slight deviation in the manipulate activation of hand-engineered and learned behaviors. This can be attributed to the fact that manipulation activations are near-zero values, and predicting values correct to decimal places will indicate overfitting. The network activations for the end-to-end approach do not show any particular pattern. The plot verifies our hypothesis that the LSE approach makes the task tractable compared to an end-to-end approach.

For the real robot experiments, using the subtask approach, the robot can pick up different objects, whereas using an end-to-end training method, the robot can only complete the block pickup which it has been trained on and fails in grasping all other objects, Fig. \ref{fig:experiments}. 
LSE approach allows us to fine-tune LSE gripper closure for a particular subtask (in this case \textit{retract}) in order to grasp different types of objects that is not possible in an end-to-end policy. This verifies that the subtask approach can generate robust behavior by fine-tuning a subset of the subtask. Refer to the attached supplementary video.

\section{CONCLUSIONS} \label{conclusion}
This work shows that a high-level task representation of human knowledge can be leveraged to decompose a pick and place task into multiple subtasks. These subtasks can be learned independently via specialized expert networks using a DRL-based policy. We present a training strategy that does not require demonstrations and is sample-efficient compared to an imitation learning-based method studied previously. 
Furthermore, we demonstrate the successful translation of policies learned in a simulated scene to the real robotic system. Using our approach, the real robotic system can grasp different geometric shapes. 
 
Future work will be focused on the decomposition of the task in a self-supervised fashion. Further, we will expand the repertoire of the subtasks that can be fused deferentially to adapt to a new task.




\section*{ACKNOWLEDGMENT}


The authors would like to acknowledge the support from the European Union's Horizon 2020 research and innovation program under the Marie Sklodowska-Curie grant agreement No 813782 (ATLAS) and under grant agreement No. 742671 (ARS). Authors would like to thank Enrico Sgarbanti for the support in real robot experiments.
\bibliographystyle{IEEEtran}
\bibliography{root}

\end{document}